\documentclass[10pt,twocolumn,letterpaper]{article}

\usepackage{cvpr}
\usepackage{times}
\usepackage{epsfig}
\usepackage{graphicx}
\usepackage{amsmath}
\usepackage{amssymb}
\usepackage[]{algorithm2e}
\usepackage{resizegather}
\usepackage{xcolor}
\renewcommand{\thefootnote}{\arabic{footnote}}	
\usepackage[pagebackref=true,breaklinks=true,letterpaper=true,colorlinks,bookmarks=false]{hyperref}

\cvprfinalcopy 

\def\cvprPaperID{24} 

\ifcvprfinal\fi
\begin{document}

\title{ Monte-Carlo Siamese Policy on Actor for Satellite Image Super Resolution}
\author{Litu Rout$^1$ \\
\and Saumyaa Shah$^2$ \and S Manthira Moorthi$^1$ \and Debajyoti Dhar$^1$ \and $^1$ Signal and Image Processing Group\\
Space Applications Centre\\
Indian Space Research Organisation\\
{\tt\small (lr, smmoorthi, deb)@sac.isro.gov.in} \and  $^2$ Work done at Space Applications Centre\\
{\tt\small saumyaashah2498@gmail.com}
}

\maketitle

\begin{abstract}
   In the past few years supervised and adversarial learning have been widely adopted in various complex computer vision tasks. It seems natural to wonder whether another branch of artificial intelligence, commonly known as Reinforcement Learning (RL) can benefit such complex vision tasks. In this study, we explore the plausible usage of RL in super resolution of remote sensing imagery. Guided by recent advances in super resolution, we propose a theoretical framework that leverages the benefits of supervised and reinforcement learning. We argue that a straightforward implementation of RL is \textit{not} adequate to address ill-posed super resolution as the action variables are not fully known. To tackle this issue, we propose to parameterize action variables by matrices, and train our policy network using Monte-Carlo sampling. We study the implications of parametric action space in a model-free environment from theoretical and empirical perspective. Furthermore, we analyze the quantitative and qualitative results on both remote sensing and non-remote sensing datasets. Based on our experiments, we report considerable improvement over state-of-the-art methods by encapsulating supervised models in a reinforcement learning framework.\let\thefootnote\relax\footnote{\noindent Accepted for publication at Computer Vision and Pattern Recognition (CVPR) Workshop on Large Scale Computer Vision for Remote Sensing Imagery.}

\end{abstract}

\section{Introduction}
 Despite significant progress in complex environments~\cite{mnih2013playing,mnih2015human,lee2018modular}, Deep Reinforcement Learning (DRL) has not received much needed attention from remote sensing community. In this study, we intend to bridge this gap by providing a DRL framework to tackle single image super-resolution in the context of satellite image processing. 
 
 Reinforcement Learning (RL) is a sequential decision making process that focuses on maximizing long-term expected return by interacting with the environment iteratively~\cite{sutton1998introduction}. In the process of maximizing reward, function approximation plays a vital role~\cite{sutton2000policy}. In the recent years, several function approximators have been proposed to estimate key ingredients of RL: action value function (Q-function) and state value function (V-function). These functions are estimated differently in two broader categories of reinforcement learning methods: model-based and model-free.
 
 Both model-based and model-free methods have their own merits and demerits. Model-based methods build a representative model of the environment and then sample rewards and transitions based on this approximation to estimate value functions. As per recent studies~\cite{levine2013variational}, model-based methods are more efficient in discrete environment due to low sample complexity. On the other hand, model-free methods learn the value functions by rollout samples obtained directly via interaction with the environment~\cite{sutton1998introduction}. While model-free methods are suitable for continuous and complex Markov Decision Processes (MDPs) in general, these methods suffer from high sample complexity~\cite{lillicrap2017continuous}. In this study, we primarily focus on model-free reinforcement learning as we do not have a model of the super-resolution environment. 
 
 Model-free control algorithms, such as Monte-Carlo (MC) and Temporal Difference (TD) have shown appealing results in numerous decision making processes~\cite{sutton1998introduction,guo2014deep}. In most of these MDPs, there are two commonly used function approximators: value based and policy based. While value based function approximation is efficient in low dimensional or discrete action space, it does not scale well to continuous action space. In addition, it is less effective in learning stochastic policies. On the contrary, the policy based approximators can learn stochastic policies in high dimensional continuous action space. However, the policy based methods typically converge to local rather than global optimum and have high variance in the estimation. 

Several policy based algorithms have been proposed over the years, such as REINFORCE~\cite{williams1992simple}, Actor-Critic~\cite{wang2016sample,haarnoja2018soft}, Proximal Policy Optimization~\cite{schulman2017proximal,schulman2015trust}, and Supervised Policy Update~\cite{vuong2018supervised}. A straightforward implementation of these algorithms is not sufficient for single image super-resolution. Most of these algorithms require adequate information about the action space in order to estimate an optimal policy. For instance, the control variables are known in most continuous action space based RL environments, though the values of these variables are estimated based on stochastic policy. Contrary to that, there are several real world control problems in which it is difficult to explicitly model the variables of action space. In such environments, we propose a way to find optimal control policy by allowing the agent to take parametric actions. To put more succinctly, the action variables are represented by matrices and the values within the matrices characterize magnitude of that action. Of particular interest, reinforcement learning based super resolution is one such environment where an actor is expected to transit from a low resolution state to a high resolution state via sequence of actions. In this environment, the sequential parametric actions are taken by shallow neural networks and a reward is received only at the end of an episode provided the terminating state falls within an $\epsilon$-ball of the high resolution state. In this process, the policy network guides the agent by providing probabilistic confidence on the performed actions in each episode. 

Many researchers have applied DRL in challenging computer vision problems~\cite{caicedo2015active,cao2017attention}. Caicedo \etal~\cite{caicedo2015active} used a set of actions as a part of sequential decision making process and rewarded the agent when the transformed bounding box had optimum overlap with target bounding box. Cao \etal~\cite{cao2017attention} exploited global inter-dependency of images to hallucinate missing high frequency details using attention-aware RL agent. Our work differs from these previous approaches in a sense that we do not use explicit set of action variables, such as move left or move right with continuous action values, i.e., the amount of movement in these directions. Instead, we use parametric action variables which allows to perform relevant actions with continuous action values. The underlying hypothesis is that a particular combination of these parameters may lead to a certain action which otherwise would not have been apparent in discrete action space. In other words, a particular combination may allow the agent to take an action that results in edge detection, or another combination may give rise to color feature extraction. Here, the action variable could be edge detection with sharpness of these edges represented by action values. Thus, our primary contribution in this study is to propose a novel reinforcement learning framework to perform the complex task of super resolution. Further, we intend to provide theoretical and empirical evidence of the proposed framework that is shown to outperform state-of-the-art methods in remote sensing. 

The rest of the paper is organized as follows. In Section~\ref{rel_work}, we briefly discuss about prior and concurrent works done to address the problem under investigation. Section~\ref{prelim} contains the preliminary settings of DRL followed by Section~\ref{method} which describes the proposed method in detail. We provide theoretical evidence in Section~\ref{theory} along with empirical experiments in Section~\ref{exps}. At the end, we draw concluding remarks and suggest future line of research in Section~\ref{conc}.

\section{Related Work}
\label{rel_work}
Recent advances in deep learning has created a surge in single image super resolution. Starting with the pioneering work of Dong \etal~\cite{dong2015image}, deep learning based super resolution has been actively explored and often outperforms state-of-the-art methods on various benchmark datasets~\cite{martin2001database,arbelaez2010contour,zeyde2010single,bevilacqua2012low,huang2015single,timofte2016seven,agustsson2017ntire,blau20182018}. Thereafter, Lai \etal~\cite{lai2018fast} proposed a deep Laplacian pyramid network for fast and accurate super resolution. It progressively upsampled the coarse resolution band to decompose the difficult task into relatively simple sub-problems. Among other supervised learning framework, Anwar \etal~\cite{anwar2019densely} proposed Densely Residual Laplacian Network (DRLN) that achieved state-of-the-art results on almost all benchmarks. Further, Ledig \etal~\cite{ledig2017photo} introduced an adversarial framework to push the reconstructed images towards natural manifold of realistic data. Wang \etal~\cite{wang2018esrgan} improved upon this idea and designed a generative model which achieved higher perceptual quality. A detailed discussion on recent developments in super resolution can be found in~\cite{wang2019deep,anwar2019deep}.

Remote sensing image super resolution is becoming increasingly popular. Particularly intriguing is the complex spatial distribution of remote sensing imagery which makes super resolution a relatively hard problem. Beyond academic interests, multi-band images are especially useful in wide variety of domains including agriculture~\cite{lacar2001use}, surveillance~\cite{uzkent2017aerial} and land cover classification~\cite{chen2015spectral}. In addition, the fundamental ideas developed in computer vision community are becoming prevalent in certain applications based on remotely sensed multi-band imagery~\cite{bastidas2019channel,dong2019transferred,ma2019super,gu2019deep,jiang2019edge}. 

After decades of research devoted in supervised and adversarial learning, it seems natural to wonder whether another branch of artificial intelligence, namely Reinforcement Learning (RL) would benefit the super resolution community. There has been little study of DRL in single image super resolution~\cite{yu2018crafting,yu2019path}. Yu \etal~\cite{yu2018crafting} have taken a step along this interesting direction of research by dynamically selecting a toolchain for progressive restoration. Further, Yu \etal~\cite{yu2019path} devised a framework by combining deep learning with REINFORCE to restore non-remote sensing images under noisy environment. In this study, we further explore the plausible usage of DRL in the context of satellite image super resolution. We provide theoretical results supported by experimental evidence to corroborate our hypothesis.

\section{Preliminaries and Notations}
\label{prelim}
A Markov Decision Process (MDP) is defined as a tuple $\left ( S,A,R,P,\gamma  \right )$, where $S$ is the continuous or discrete state space, $A$ is the continuous or discrete action space, $R$ is the immediate reward function, $P$ is the transition probability, and $\gamma \in (0,1)$ is the discount factor. The goal of an agent is to find an optimal policy $\pi^{*}$ that maximizes its expected reward,
\begin{equation}
    \pi^{*} = \arg \max_{\pi \in \Pi} \mathcal{J}\left (\pi \right),
\end{equation}
where $\Pi$ is the set of policies and $\mathcal{J}\left (\pi \right)$ is the policy evaluation metric defined by,
\begin{equation}
    \mathcal{J}\left ( \pi \right ) = \mathbb{E}_{\tau \in \pi } \left [ \sum_{t=1}^{T+1} \gamma^{t-1} r_{t}  \right ].
\end{equation}
Here, $T$ represents the time step of terminal state in each episode and $\tau$ is the trajectory, $(s_0, a_0, r_1, s_1, a_1, r_2,\dots, s_T,a_T,r_{T+1})$ sampled from policy $\pi$. As we are focusing on model-free approaches, the state and action value functions (V and Q) are approximated based on sampled trajectories unlike model-based approaches where full width backup is taken into consideration as the transition dynamics is accessible. In common policy optimization strategy, the policy $\pi$ is parameterized by $\theta$ where the objective is to find optimal set of parameters $\theta^{*}$ that maximizes expected reward,
\begin{equation}
    \theta^{*} = \arg \max_{\theta} \mathcal{J}\left ( \theta \right),
\end{equation}
\begin{equation}
     \mathcal{J}\left ( \theta \right ) = \sum_{s \in S} d^{\pi_\theta}\left ( s \right )\sum_{a \in A} \pi_\theta\left ( s,a \right ) R_{s,a},
\end{equation}
where $d^{\pi_\theta}\left ( s \right )$ is a stationary distribution of Markov chain for $\pi_\theta$ and $R_{s,a}$ is the reward function for state $s$ and action $a$. The policy parameters are updated by $\theta \leftarrow \theta + \Delta \theta$, where $\Delta \theta$ is computed by the famous likelihood trick~\cite{williams1992simple},
\begin{equation}
  \Delta \theta = \nabla_{\theta} \mathcal{J}\left ( \theta \right ) = \mathbb{E}\left [ R_{s,a} \nabla_\theta \log \pi_\theta \left ( s,a \right ) \right ].
  \label{del_theta}
\end{equation}
There are several variants of equation~(\ref{del_theta}) that emphasize on faster convergence to optimal solution and robust policy estimation. To study and analyze the proposed idea at a fundamental level, we choose a simple and effective policy gradient strategy, namely MC-REINFORCE~\cite{williams1992simple} as given in equation~(\ref{del_theta}). However, the proposed approach is not limited to MC-REINFORCE, and would certainly benefit from recent advances in policy optimization.

\section{Methodology}
\label{method}
The idea of parameterizing action space/variables is inspired by the notion of building a model of the environment in model-based RL. In a model-based RL, the transition dynamics $(P)$ and reward function $(R)$ are parameterized assuming that the state space $(s)$ and action space $(a)$ are known. The proposed approach is slightly different from this model-based approach in a sense that we parameterize the action space $(a)$ and learn the policy in a model-free way using MC sampling.

\subsection{Representation Learning}
\label{rl}
 Here, we discuss about efficient representation of each state in our MDP as it plays a vital role in solving MDPs~\cite{sutton1998introduction}. Instead of naively representing each state, we use Convolutional Neural Network (CNN) as feature extractor due to its tremendous success in learning latent representation. The feature extractor network, $\Phi(s)$ parameterized by $\theta_{f}$ operates on each state, $s \in \mathbb{R}^{H\times W \times C}$, 
 \begin{equation}
     \Tilde{s} = \Phi(s ;\theta_f),~\Tilde{s} \in \mathbb{R}^{H\times W \times \Tilde{C}},
 \end{equation}
 where $H,~W,~C,~\text{and}~\Tilde{C}$ represent height, width, input channels, and number of feature maps, respectively. The output of neural network, $\Phi(s ;\theta_f)$ is computed by, 
 \begin{equation}
      \Phi(s ;\theta_f) := FE_n\left (FE_{n-1}\left ( \dots \left ( FE_0\left ( s \right ) \right ) \right )  \right ),
 \end{equation}
 where $FE$ represents Feature Extraction block consisting of one convolution and one LeakyReLU unit. Here, $n$ represents number of FE blocks.

\subsection{Actor Network}
\label{an}
The Actor Network (AN), $\Omega_{\theta_{a}}(.)$ parameterized by $\theta_{a }$ performs parametric actions on the latent representation of state space, $\Tilde{s}$. Each action is parameterized by a shallow neural network consisting of a single Residual Block (RB). To span the dynamic range of each state, we use a customized RB, as given by equation~(\ref{eq_rb}), in contrast to the one proposed in~\cite{he2016deep}.
\begin{equation}
\label{eq_rb}
    RB(x) = x+\lambda h(x),
\end{equation}
where $h(x)$ is a sequential neural network consisting of \{convolution, ReLU, and convolution\} units. Here, $\lambda$ represents the scaling factor. The agent performs sequence of actions, $a^{RB}_{n}(.)$ and the intermediate states are computed by,
\begin{equation}
    \Tilde{s}_{n} = a^{RB}_{n}(\Tilde{s}_{n-1}),~ n={1,2,\dots,N},
\end{equation}
where N represents total number of action variables in our MDP. Here, $\Tilde{s}_0$, $\Tilde{s}_n$, and $\Tilde{s}_N$ represent the latent representation of the initial, intermediate, and arrived state, respectively. Different combinations of these parameters present in each kernel of these action variables lead to different actions necessary to achieve the desired goal. The latent representation of arrived state, $\Tilde{s}_N$ is passed through a cascade of Transition Blocks $(TB)$ in order to map latent space into state space, as given in equation~(\ref{eq_tb}).
\begin{equation}
    \hat{s} = TB_m\left ( TB_{m-1}\left ( \dots \left ( TB_0\left ( \tilde{s}_{N} \right ) \right ) \right ) \right )
    \label{eq_tb}
\end{equation}
Here, $\hat{s} \in \mathbb{R}^{H \times W \times C}$, and each $TB$ consists of one convolution and one LeakyReLU unit. Since the agent receives reward at the end of each episode, we only convert the final latent representation, $\Tilde{s}_N$ to state space, $\hat{s}$ for minimizing time complexity. 

\subsection{Siamese Policy Network}
\label{spn}
Here, we provide a justification for parameterizing policy network using Siamese architecture. The standard policy network, $\pi_{\theta_{p}}(s,a)$ parameterized by $\theta_{p}$ provides a distribution over actions, $a$ given a state, $s$. Thus, the policy network provides a probabilistic view of how good it is to take an action at a given state. In other words, it imposes a confidence on the agent's actions at a particular state. If the sequence of actions triggers a transition such that the final state falls within an $\epsilon$-ball of the goal state, then the confidence level on agent's actions is enhanced. In such scenarios, the policy gradient approach increases the likelihood of taking these relevant actions.

To estimate the confidence on agent's actions, we propose to use Siamese neural network architecture~\cite{bromley1994signature}. The Siamese Policy Network (SPN), $\Psi_{\theta_{p}}(\hat{s},s^{*})$ measures the discrepancy between arrived state, $\hat{s}$ and goal state, $s^{*}$. The SPN is stochastic in nature due to which it does not require the environment to be noisy in order to perform sufficient exploration. The two branches of Siamese neural network take $\hat{s}$ and $s^{*}$ as inputs, projects them into feature space using shared parameters across both branches, and correlate them in feature space to better estimate their discrepancy,
\begin{equation}
     \Psi_{\theta_p}(\hat{s},s^{*}) = \Phi_{\theta_p}(\hat{s}) * \Phi_{\theta_p}(s^{*}) + b,
\end{equation}
where $b \in \mathbb{R}$ and $\Phi_{\theta_p}(.)$ represents the CNN in each branch with shared parameters $\theta_p$. The probabilistic confidence is then computed by sigmoidal activation unit,
\begin{equation}
    \pi_{\theta_p}\left ( s,a \right ) = \frac{1}{1+\exp(-\Psi_{\theta_p}(\hat{s},s^{*}))}.
\end{equation}
The arrived state ($\hat{s}$) is a function of implicit actions, $a$ containing $a_n^{RB}, n=1,2,\dots N$. In the super-resolution environment, the model receives reward only at the final state based on its Euclidean distance from target state.

\subsection{Siamese Policy On Actor}
\label{spoa}
The proposed method, which we call Siamese Policy On Actor (SPOA), encapsulates representation learning, AN, and SPN to provide an end to end DRL framework for image super resolution. Motivated by the findings of Goodfellow \etal~\cite{goodfellow2014generative}, we propose to allow two networks, namely actor and policy to supplement each other in the learning process so as to find a global optimum. 

In the current setting, we consider occurrence of each observable state to be equally likely, i.e., $d^{\pi_\theta} (s) \sim \mathbb{U}$, where $\mathbb{U}$ denotes uniform distribution. We define reward function, $R_{s,a}$ as negative mean squared error between $\hat{s}$ and $s^{*}$. The agent receives reward at the end of an episode and it is maximum when $\left \| \hat{s}-s^{*} \right \|^2$ falls within an $\epsilon$-ball around $s^{*}$. We use $n=3$ FE blocks in representation learning, $N=3$ RBs, $m=3$ TBs in AN, and 3 blocks of \{convolution, LeakyReLU\} units in SPN.

\section{Theoretical Results}
\label{theory}
In this section, we elaborate on the supplementary training procedure. We independently train AN and SPN in a least expensive way before training SPOA. Thus, we ensure that the arrived state does not reside far away from initial state, which otherwise would make it unattainable.

\textit{\textbf{Lemma I}: Training AN} \\
Let $\theta_{fa} = \left \{ \theta_{f}, \theta_{a} \right \}$ and $\mathcal{J}(\theta_{fa})$ denote the expected return accumulated by the agent with a given policy $\pi_{\theta_p}$,
\begin{equation}
     \mathcal{J}(\theta_{fa}) = \mathbb{E}\left [ R_{s,a} \right ] = \mathbb{E}\left [ -(\hat{s}-s^{*})^{2} \right ]. 
\end{equation}
The parameters are updated by, $\theta_{fa} \leftarrow \theta_{fa}+\Delta \theta_{fa}$ where,
\begin{equation}
\label{fa}
 \Delta \theta_{fa} = \mathbb{E}\left [ -2\left ( \hat{s}-s^{*} \right )~\nabla_{\theta_{fa}} \left ( TB_{[m]}\left ( \Omega_{\theta_a} \left ( \Phi_{\theta_f}\left ( s \right ) \right ) \right ) \right ) \right ].
\end{equation}
Here, $[m]$ represents a set of $\left \{ 0,1,\dots,m \right \}$.

By stochastic gradient ascent, the update equation~(\ref{fa}) becomes
\begin{equation}
\label{fa1}
  \Delta \theta_{fa} = -\alpha\left ( \hat{s} - s^{*} \right )\nabla_{\theta_{fa}}\left ( TB_{[m]}\left ( \Omega_{\theta_a} \left ( \Phi_{\theta_f}\left ( s \right ) \right ) \right ) \right ),
\end{equation}
where $\alpha$ denotes step size.

\textit{\textbf{Lemma II}: Training SPN}\\
Let $\mathcal{J}(\theta_{p})$ denotes the expected return accumulated by the agent with fixed set of parameters ( $\theta_{fa}$), then
\begin{equation}
     \mathcal{J}\left ( \theta_p \right ) = \mathbb{E}_{\theta_p} \left [ r \right ] = \sum_{s \in S} d^{\pi_\theta}\left ( s \right )\sum_{a \in A} \pi_\theta\left ( s,a \right ) R_{s,a}.
\end{equation}
Using stochastic gradient ascent, the parameters are updated using the famous likelihood trick~\cite{sutton1998introduction} as given by
\begin{equation}
\label{p}
    \theta_{p} \leftarrow \theta_{p}+\Delta \theta_{p},~
    \Delta \theta_{p} = \nabla_{\theta_p}\mathcal{J}(\theta_p)=\beta R_{s,a}\nabla_{\theta_p} \log \pi_{\theta}\left ( s,a \right ),
\end{equation}
where $\beta$ denotes step size.

\textit{\textbf{Theorem I}: Training SPOA}\\
\textit{Let $\theta = \left \{ \theta_f, \theta_a, \theta_p \right \}$ and $\mathcal{J}(\theta)$ denotes the expected return. The parameters of SPOA ($\theta$) are updated by $\theta \leftarrow \theta + \Delta \theta$ where,
\begin{equation}
    \Delta \theta = \Delta \theta_{p} + \Delta \theta_{fa}.
\end{equation}
}
\textit{\textbf{Proof:}}
\begin{equation*}
    \begin{split}
        \mathcal{J}(\theta) & = \mathbb{E}\left [ r \right ] = \sum_{s \in S} d^{\pi_{\theta_p}}\left ( s \right ) \sum_{a \in A}\pi_{\theta_p}\left ( s,a \right ) R_{s,a}\\
        \Delta \theta & = \nabla_\theta \mathcal{J}(\theta) \\
        & = \sum_{s \in S} d^{\pi_{\theta_p}}\left ( s \right ) \sum_{a \in A} \nabla_\theta \left ( \pi_{\theta_p}\left ( s,a \right ) R_{s,a} \right )\\
        & = \sum_{s \in S} d^{\pi_{\theta_p}}\left ( s \right ) \sum_{a \in A} \left ( R_{s,a} \nabla_\theta  \pi_{\theta_p}\left ( s,a \right )  +  \pi_{\theta_p}\left ( s,a \right ) \nabla_\theta R_{s,a}  \right )\\
        & = \sum_{s \in S} d^{\pi_{\theta_p}}\left ( s \right ) \sum_{a \in A} R_{s,a} \nabla_{\theta_p} \pi_{\theta_p}\left ( s,a \right ) \\ &\mathrel{\phantom{=}} + \sum_{s \in S} d^{\pi_{\theta_p}}\left ( s \right ) \sum_{a \in A}  \pi_{\theta_p}\left ( s,a \right ) \nabla_{\theta_{fa}}  R_{s,a} \\
        & = \sum_{s \in S} d^{\pi_{\theta_p}}\left ( s \right ) \sum_{a \in A} R_{s,a} \pi_{\theta_p}\left ( s,a \right ) \nabla_{\theta_p} \log \pi_{\theta_p}\left ( s,a \right ) \\ &\mathrel{\phantom{=}} + \sum_{s \in S} d^{\pi_{\theta_p}}\left ( s \right ) \sum_{a \in A}  \pi_{\theta_p}\left ( s,a \right ) \nabla_{\theta_{fa}}  R_{s,a} \\
        & = \mathbb{E}\left [ R_{s,a} \nabla_{\theta_p}\log \pi_{\theta_p}\left ( s,a \right ) \right ]+\mathbb{E}\left [ \nabla_{\theta_{fa}}R_{s,a} \right ]\\
    \end{split}
\end{equation*}

Using stochastic gradient ascent, $\Delta \theta$ becomes,
\begin{equation}
\label{prf}
\begin{split}
    \Delta \theta & = \beta R_{s,a}\nabla_{\theta_p}\log \pi_{\theta_p}\left ( s,a \right ) \\ & - \alpha \left ( \hat{s} - s^{*} \right )\nabla_{\theta_{fa}}\left ( TB_{[m]}\left ( \Psi_{\theta_a} \left ( \Phi_{\theta_f}\left ( s \right ) \right ) \right ) \right ).
\end{split}
\end{equation}
From \textit{\textbf{Lemma I}} and \textit{\textbf{Lemma II}}, equation~(\ref{prf}) becomes,
\begin{equation}
    \Delta \theta = \Delta \theta_{p} + \Delta \theta_{fa}.
\end{equation}

The pseudo code of SPOA is given in \textbf{Algorithm} ~\ref{algo}. We sample batches of experiences and use replay buffer to update the SPOA parameters for the entire batch. \\

\begin{algorithm}[!t]
\caption{Monte-Carlo Siamese Policy On Actor}
 \label{algo}
\SetAlgoLined
\KwResult{SPOA parameters, $\theta$}
 initialize $\theta$\;
 
 \For{episode = 1,2,...,E}{
  initialize empty replay buffer $\mathbb{D}$\;
  \While{$\mathbb{D}$ not full}{
 	Sample initial state, $s_0 \sim \mathbb{U}$\;
 	Sample corresponding goal state, $s^{*}$\;
 	}
  
  \For{actor=1,2,...,A}{
  Take parametric sequential actions on $\mathbb{D}$\;
  Compute $\Delta \theta_{fa} = -\alpha\left ( \hat{s} - s^{*} \right )\nabla_{\theta_{fa}}\left ( TB_{[m]}\left ( \Omega_{\theta_a} \left ( \Phi_{\theta_f}\left ( s \right ) \right ) \right ) \right )$\;
  Update $\theta_{fa} \leftarrow \theta_{fa}+\Delta \theta_{fa}$\;
  }
  \For{policy=1,2,...,P}{
  Given actor parameters $\theta_{fa}$, follow parametric policy $\pi_{\theta_p}(s,a)$ on $\mathbb{D}$\; Compute $\Delta \theta_{p} = \beta R_{s,a}\nabla_{\theta_p} \log \pi_{\theta}\left ( s,a \right )$\;
  Update $\theta_{p} \leftarrow \theta_{p}+\Delta \theta_{p}$\;
  }
  \For{spoa=1,2,...,S}{
  Follow policy with implicit actions on $\mathbb{D}$\;
  Compute new $\Delta \theta_{p}$ and $\Delta \theta_{fa}$\;
  Compute $\Delta \theta = \Delta \theta_{p} + \Delta \theta_{fa}$\;
  Update  $\theta \leftarrow \theta+\Delta \theta$\;
  }
}
\end{algorithm}

\section{Experiments}
\label{exps}

\subsection{Datasets and Study Area}
Here, we describe the datasets used to analyze the performance of the proposed methodology in two folds. First, we develop the theoretical foundation, and validate the pipeline experimentally on 1000 images of CelebA~\cite{liu2018large} and 3000 patches (64x64) of Indian Remote Sensing satellite (IRS-1C). While these datasets help us verify the efficacy of proposed theoretical formulation, it is hard to infer the generalization ability from them. For this reason, we extend our analysis to large scale remote sensing images of WorldView-2. With 80-20 split we use 40000 patches of WorldView-2 over Washington having Ground Sampling Distance (GSD) 1.84m.

\subsection{Implementation Details}
\textbf{Data Preparation:} In this study, the scaling factor between Low Resolution (LR) and High Resolution (HR) images is set to 4x. To prepare training data, we crop 64x64 patches from the training HR images. Following Dong \etal~\cite{dong2015image}, the LR training patches are obtained by downsampling the HR patches by a factor of 4 using bicubic kernel. For data augmentation, we randomly choose one of the following techniques: rotation by 90 degree, horizontal flips or vertical flips.

\textbf{Network Architecture:} During development stage, we use SRCNN~\cite{dong2015image} as the backbone of the actor network to establish the theoretical foundation. To assimilate the performance of SPOA built upon deeper architectures, we explore various state-of-the-art methods. Motivated by recent advances, we use DRLN~\cite{anwar2019densely} with a network depth of 4 cascading residual-in-residual blocks in our final AN. For the policy network, we use convolutional layers with kernel size (3,3), followed by LeakyReLU activation with a negative slope of 0.1. All trainable parameters are initialized using Xavier method~\cite{glorot2010understanding}.

\textbf{Training Details:} 
The interpolated image and the corresponding high resolution image represent the initial and goal state in our MDP, respectively. In each episode, we sample from uniformly distributed initial state space. We set total episodes to 100000 and replay buffer size, $\mathbb{D}$ = 10. For training, we use different step sizes, i.e., $\alpha = 1e-4~\&~\beta=1e-7$ and Adam optimizer~\cite{kingma2014adam}. A least expensive solution is chosen to update AN, SPN, and SPOA, i.e, $A=1$, $P=1$, and $S=1$. We use a common system configuration with 2x Tesla K40 in all our experiments. The SPOA learning algorithm has been implemented with PyTorch~\cite{paszke2017automatic}.

\textbf{Evaluation metrics:} The resultant super resolved images are evaluated using four commonly used image quality metrics: PSNR~\cite{hore2010image}, SSIM~\cite{hore2010image}, SRE~\cite{lanaras2018super}, and SAM~\cite{yuhas1992discrimination}. However, according to Ledig \etal~\cite{ledig2017photo}, these distortion measures fundamentally go against human perception of image quality. For this reason, to assimilate perceptual quality, we use no reference measures like NIQE~\cite{mittal2012making} and Ma's score~\cite{ma2017learning}.  Using these measures, we compute the Perceptual Index (PI) of an image as specified in the PIRM-SR challenge~\cite{blau20182018}.

\subsection{Analysis on CelebA}
 \begin{figure}[t]
     \centering
     \includegraphics[width=\columnwidth]{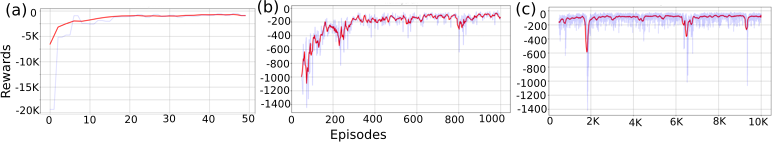}
     \caption{Learning dynamics. The blue curve shows actual reward gathered per episode. The red curve shows windowed average of actual reward per episode. We use a forward window of size 10.}
     \label{learn}
 \end{figure}
Figure~\ref{learn} shows gradual development in accumulating rewards over multiple episodes. The agent initially performs random actions, which are sampled from the proposed stochastic SPN, in the form of exploration. This is evident from Figure~\ref{learn}(a), where the accumulated reward is not so surprisingly very low in the earlier episodes. The agent however discovers relevant actions as the interaction with the environment progresses and simultaneously, SPOA increases the likelihood of these particular actions. The agent, therefore, observes a steady growth in gathering rewards, as shown in Figure~\ref{learn}(b). Once the agent figures out relevant actions, it repeatedly performs those actions in order to accumulate maximum rewards. Thereby, it reaches in the proximity of goal state from most of the initial states in almost every episodes, as shown in Figure~\ref{learn}(c).

 \begin{figure}[t]
     \centering
     \includegraphics[width=\columnwidth]{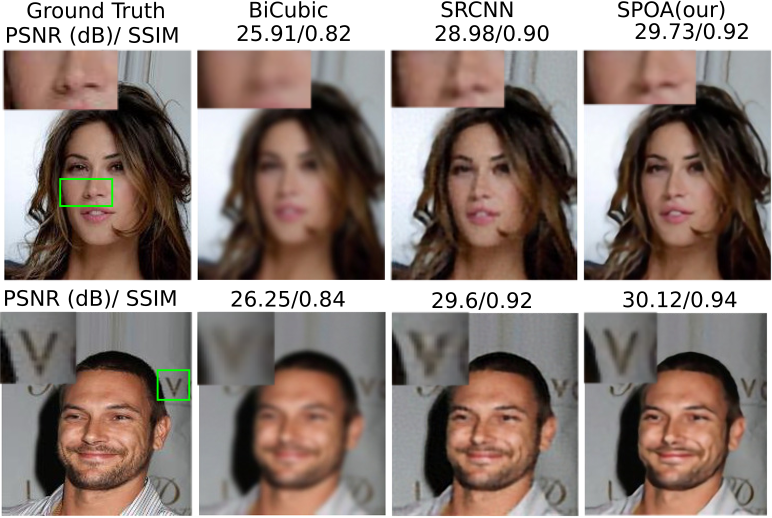}
     \caption{Qualitative analysis on CelebA. Comparison with existing high resolution data. The proposed DRL based approach, SPOA performs favourably against compared approaches.} 
     \label{pr1x}
 \end{figure}
 
 Further, we compare SPOA with BiCubic and SRCNN~\cite{dong2015image} in both training and testing datasets. It is worth mentioning that both SRCNN and SPOA share similar architecture to assert direct comparison between these two learning algorithms. Nevertheless, one can implement more sophisticated architectures in the proposed framework to gain optimal benefits. In Figure~\ref{pr1x}, we compare the reconstructed images of BiCubic, SRCNN and SPOA both qualitatively and quantitatively using PSNR (dB) and SSIM. As per the analysis, the reconstructed image by SPOA has higher structural similarity with existing high resolution data and also, it contains relatively less noise in each pixel. Further, this shows the efficacy of hierarchical composition of local constituent functions, such as implicit actor networks to learn a compact continuous mapping.

\subsection{Analysis on IRS-1C}
 \begin{figure}[t]
     \centering
     \includegraphics[width=\columnwidth]{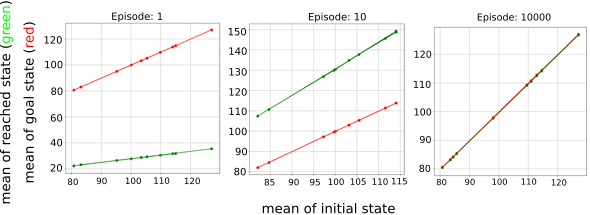}
     \caption{Mean state transition analysis. The implicit actor reaches goal state from almost every initial state.} 
     \label{dynamics}
 \end{figure}
 
\begin{figure}[t]
     \centering
     \includegraphics[width=\columnwidth]{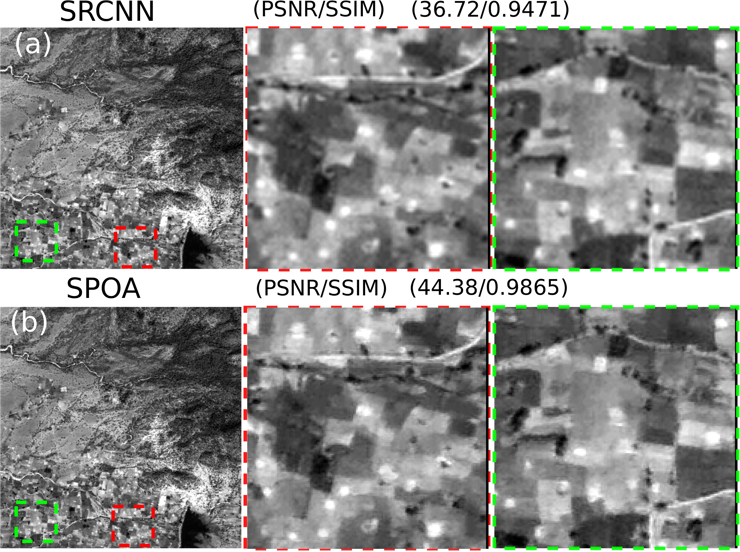}
     \caption{Qualitative analysis on IRS-1C. SPOA performs reasonably well on IRS-1C imagery.} 
     \label{irs}
 \end{figure}

Here, we analyze the performance of SPOA on IRS-1C imagery. As shown in Figure~\ref{dynamics}, the distance between initial state and goal state gradually decreases as the training progresses. Finally, the implicit actor reaches at the corresponding goal state starting from almost every initial state. In addition, Figure~\ref{irs} shows consistent improvement of SPOA over SRCNN both qualitatively and quantitatively. 

\subsection{Comparison with State-of-the-art}
To gain further insight about generalization ability of SPOA, we start our discussion by comparing the proposed framework with state-of-the-art methods. It is to be noted that we use SRCNN in SPOA for the sole purpose of building overall pipeline. However, our final framework is built upon DRLN~\cite{anwar2019densely}. To discern the usefulness of the proposed method, we study its performance on large scale remote sensing imagery of WorldView-2. In this regard, we analyze the super-resolved images in terms of both perception and distortion metrics~\cite{blau2018perception}. As given in Table~\ref{sota_tbl}, DRLN achieves state-of-the-art result among prior approaches on WorldView-2. While the proposed SPOA(DRLN) performs sub-optimally in terms of distortion metrics, it achieves higher perceptual quality in terms of NIQE, Ma's score, and PI. Consistent with the theoretical justification of Blau \etal~\cite{blau2018perception}, we observe perception-distortion tradeoff similar to adversarial networks~\cite{wang2018esrgan}. Moreover, the perception metric values of SPOA(DRLN) are relatively closer to Ground Truth (GT) as compared to state-of-the-art methods. Since SPOA derives its foundation from reinforcement learning paradigm, which is quite different from adversarial learning, it will certainly be interesting to study the theoretical basis of such similarity in perception-distortion tradeoff~\cite{blau2018perception}.

\begin{table}[t]
\centering
\resizebox{\columnwidth}{!}{%
\begin{tabular}{|l|l|l|l|l|l|l|l|}
\hline
Metrics    & PSNR                         & SSIM                         & SRE                          & SAM                          & NIQE                          & Ma's                          & PI                            \\ \hline \hline
BiCubic    & 57.51                        & 0.9939                        & 46.48                        & 17.25                        & 5.50                        & 3.77                         & 5.86                        \\
SRCNN~\cite{dong2015image}      & 59.15                        &  0.9964 & 48.10 & 14.14                        & 5.73                       & 4.88                        & 5.42                        \\
LapSRN~\cite{lai2018fast}     & 59.31                        & 0.9964 & 48.08                        & 13.98                        & 5.08                       & 5.96                        & 4.56                        \\
DRLN~\cite{anwar2019densely}       & 59.32 & 0.9964 & 48.10 &  13.97 & 4.21                        & 6.03                        & 4.08                        \\ \hline \hline
SPOA(DRLN) & 58.89                        &  0.9960 & 47.94                        & 14.69                        & {\color[HTML]{CB0000} 3.65} & {\color[HTML]{CB0000} 6.60} & {\color[HTML]{CB0000} 3.52} \\
SPOA(DRLN)+SA & {\color[HTML]{CB0000} 59.33} & {\color[HTML]{CB0000} 0.9966} & {\color[HTML]{CB0000} 48.20} & {\color[HTML]{CB0000} 13.81} & 5.02                          & 5.54                          & 4.74                          \\
SPOA(DRLN)+SA+VGG       & 59.22 & 0.9963 & 48.23 &  14.13 & 4.30                        & 6.20                        & 4.05                        \\
SPOA(DRLN)+VGG        & 58.98 & 0.9961 & 47.94 &  14.60 & 4.16                        & 6.56                        & 3.80                       \\ \hline \hline
GT         & -                            & -                            & -                            & -                            & 2.05                        & 7.01                        & 2.52   \\ \hline                    
\end{tabular}%
}
\caption{Comparison with state-of-the-art methods.}
\label{sota_tbl}
\end{table}

\subsubsection{Ablation Study}
In addition, we explore several variants of SPOA to gain intuition about its ability to achieve better distortion quality. We start our discussion by incorporating Self-Attention (SA) units~\cite{zhang2018self} in SPOA. Attending to relevant parts of an image is an interesting line of research. Recent study shows significant improvement in image quality due to attention mechanisms~\cite{chen2016attention,zhang2018self,emami2020spa}. For this reason, we augment SPOA by adding self-attention units that work in tandem with existing Laplacian channel attention units. As given in Table~\ref{sota_tbl}, SPOA(DRLN)+SA outperforms DRLN in terms of distortion metrics. Furthermore, we study whether addition of VGG loss~\cite{ledig2017photo} results in better perceptual quality. Though introduction of VGG loss does not boost performance beyond SPOA, it certainly improves the perceptual quality of SPOA(DRLN)+SA.

\subsubsection{Analysis on WorldView-2}
\begin{figure*}[t]
     \centering
     \includegraphics[width=0.82\textwidth]{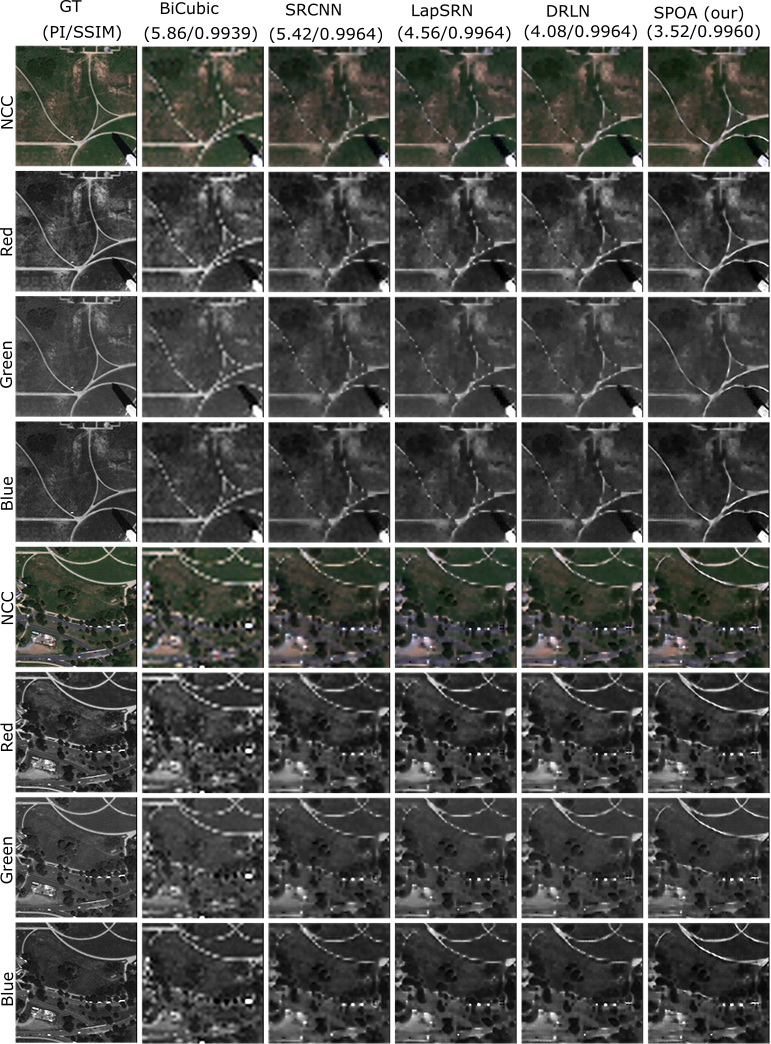}
     \caption{Qualitative analysis on WorldView-2. SPOA outperforms compared methods in perceptual quality, and also generates more natural textures while mitigating unpleasant artifacts, e.g., discontinuity of linear features.} 
     \label{sota_qual}
 \end{figure*}

To this end, we studied quantitatively how reinforcement learning driven SPOA benefits super resolution. Here, we discuss further by correlating the perception-distortion metrics with qualitative measures. As can be inferred from the Natural Color Composite (NCC) and individual bands in Figure~\ref{sota_qual}, SPOA outperforms state-of-the-art methods in terms of perceptual quality. While compared methods lack continuity in linear features, SPOA seems to preserve continuity reasonably well. Even though SSIM values are comparable, evidently the quality of super resolved images is not at par with each other. This is consistent with the observation of Blau \etal~\cite{blau2018perception}. It can be observed from Figure~\ref{sota_qual} that the sharpness and continuity of features are prominent in SPOA. In addition, SPOA produces more high frequency details which tend to improve the naturalness of super resolved images while other methods~\cite{lai2018fast,anwar2019densely} either fail to capture these details or introduce unwanted artifacts. From the individual bands of various signatures in Figure~\ref{sota_qual}, one can obtain a better visual assessment of image quality.

\section{Concluding Remarks and Future Scope}
\label{conc}
In this study, we explored the plausible usage of reinforcement learning to address complex supervised learning problems. We designed a DRL based Monte-Carlo policy gradient approach to solve model-free MDPs where adequate information about action variables is not discernible. Guided by our theoretical justification, we introduced a Siamese policy network with implicit action space. Further, we demonstrated the efficacy of the proposed method in a super resolution environment where action variables are not apparent. Using both remote sensing and non-remote sensing imagery, we studied the perception-distortion tradeoff. To satisfy the requirement on multitude of tasks, we introduced two methods: one that achieved state-of-the-art results in distortion and another, in visual perceptual quality.

A few noteworthy extensions of this paper are as follows:
\begin{enumerate}
\item \textit{Extension} of SPOA to wide variety of problems currently solved using supervised learning.

\item Instead of building upon MC-REINFORCE, one can explore the broad \textit{spectrum of reinforcement learning} algorithms in this framework.

\item Further, one can study how well SPOA figures out matrix representation of actions by \textit{hiding} known action variables in RL benchmarks.

\end{enumerate}

This study demonstrated the plausibility of DRL in solving supervised problems as sequential decision making processes. The efficacy of SPOA in this regard broadens the horizon of DRL, suggesting further investigation in this viable research direction might prove beneficial.

{\small
\bibliographystyle{ieee_fullname}
\bibliography{egbib}
}

\end{document}